\documentclass{article}

\PassOptionsToPackage{numbers, compress}{natbib}
\usepackage[final]{neurips_2024}
\usepackage{algorithm}
\usepackage{algpseudocode}
\usepackage[utf8]{inputenc} % allow utf-8 input
\usepackage[T1]{fontenc}    % use 8-bit T1 fonts
\usepackage{xcolor}     
\definecolor{darkblue}{rgb}{0, 0.17, 0.55}
\definecolor{darkgreen}{rgb}{0, 0.55, 0.12}
\definecolor{darkred}{rgb}{0.6,0,0}
\definecolor{darkgreen}{rgb}{0,0.6,0}
\usepackage[hidelinks,
              colorlinks,
              urlcolor  = darkblue,
              citecolor = darkblue,]{hyperref}       % hyperlinks
\usepackage{url}            % simple URL typesetting
\usepackage{booktabs}       % professional-quality tables
\usepackage{amsfonts}       % blackboard math symbols
\usepackage{nicefrac}       % compact symbols for 1/2, etc.
\usepackage{microtype}      % microtypography
\usepackage{xcolor}         % colors
\usepackage{colortbl}
\usepackage{graphicx}
\usepackage{subfigure}
\usepackage{wrapfig}
\usepackage{tabulary}
\usepackage{bm}
\usepackage{multirow}
\usepackage{multicol}
\usepackage{etoolbox}
\usepackage{amsmath}
\usepackage{amssymb}
\usepackage{mathtools}
\usepackage{amsthm}
\usepackage[capitalize,noabbrev]{cleveref}
\usepackage[textsize=tiny]{todonotes}

\definecolor{mycolor_blue}{HTML}{E7EFFA}
\definecolor{mycolor_green}{HTML}{E6F8E0}
\definecolor{mycolor_gray}{HTML}{ECECEC}
\definecolor{pearDark}{HTML}{2980B9}
\definecolor{mygray}{gray}{.9}
\definecolor{mycolor}{RGB}{0, 20, 205}

\theoremstyle{plain}

\theoremstyle{definition}

\theoremstyle{remark}

\title{RealCompo: Balancing Realism and Compositionality Improves Text-to-Image Diffusion Models}

\author{Xinchen Zhang$^{1*}$\ \  \quad Ling Yang$^{2}$\thanks{Contributed equally.} \ \thanks{Corresponding authors: yangling0818@163.com, bin.cui@pku.edu.cn.}
\quad Yaqi Cai$^{3}$ \quad Zhaochen Yu$^{2}$ \quad Kai-Ni Wang$^{4}$ \quad\\ \quad \textbf{Jiake Xie}$^{5}$ \quad\textbf{Ye Tian}$^{2}$\quad \textbf{Minkai Xu}$^{6}$\quad \textbf{Yong Tang}$^{5}$\quad \textbf{Yujiu Yang}$^{1}$\quad \textbf{Bin Cui}$^{2}$ \\
$^{1}$Tsinghua University \quad $^{2}$ Peking University \quad $^{3}$ University of Science and Technology of China\quad\\ \quad$^{4}$ Southeast University \quad$^{5}$ PicUp.AI\quad$^{6}$ Stanford University\\
\url{https://github.com/YangLing0818/RealCompo}
}
\begin{document}

\maketitle

\begin{abstract}
  Diffusion models have achieved remarkable advancements in text-to-image generation. However, existing models still have many difficulties when faced with multiple-object compositional generation. In this paper, we propose \textbf{\textit{RealCompo}}, a new \textit{training-free} and \textit{transferred-friendly} text-to-image generation framework, which aims to leverage the respective advantages of text-to-image models and spatial-aware image diffusion models (e.g., layout, keypoints and segmentation maps) to enhance both realism and compositionality of the generated images. An intuitive and novel \textit{balancer} is proposed to dynamically balance the strengths of the two models in denoising process, allowing plug-and-play use of any model without extra training. Extensive experiments show that our RealCompo consistently outperforms state-of-the-art text-to-image models and spatial-aware image diffusion models in multiple-object compositional generation while keeping satisfactory realism and compositionality of the generated images. Notably, our RealCompo can be seamlessly extended with a wide range of spatial-aware image diffusion models and stylized diffusion models.
\end{abstract}

\section{Introduction}

The field of diffusion models has witnessed exciting developments and significant advancements recently\citep{yang2024mastering,song2020score, ho2020denoising, song2020denoising, rassin2024linguistic}. Among various generative tasks, text-to-image (T2I) generation \citep{nichol2022glide, hu2024instruct, yang2024improving} has gained considerable interest within the community. T2I diffusion models such as Stable Diffusion \citep{rombach2022high}, Imagen \citep{saharia2022photorealistic} and DALL-E 2/3 \citep{ramesh2022hierarchical, betker2023improving} have exhibited powerful capabilities in generating images with high aesthetic quality and realism \citep{betker2023improving, podell2023sdxl}. However, they often struggle to align accurately with the compositional prompt when it involves multiple objects or complex relationships \citep{lian2023llm, bar2023multidiffusion, park2024energy}, which requires the model to have strong spatial-aware ability.

One potential solution to optimize the compositionality of generated images is providing a spatial-aware condition to control diffusion models \citep{fan2023frido, yang2023reco, wu2023self}, such as layout/boxes \citep{phung2023grounded,feng2024layoutgpt}, keypoint/pose \citep{zhang2023adding} and segmentation map \citep{huang2023collaborative}. These spatial-aware conditions are fundamentally similar in functioning, thus we mainly focus our analysis on layout-to-image (L2I) models for simplicity. With the control of layout, L2I models \citep{li2023gligen, chen2024training, xie2023boxdiff} improve compositionality by generating objects at specified locations. For instance, GLIGEN \citep{li2023gligen} designs trainable gated self-attention layers to incorporate layout input and controls the strength of its incorporation by changing parameter $\beta$. Although L2I models improve the weaknesses of compositional text-to-image generation, their generated images exhibit a significant decline in realism compared to T2I models \citep{li2023gligen, zhou2024migc}.

\begin{figure*}[t!]
\vskip -0.1in\label{fig:motivation}
\begin{center}
\centerline{\includegraphics[width=1\textwidth]{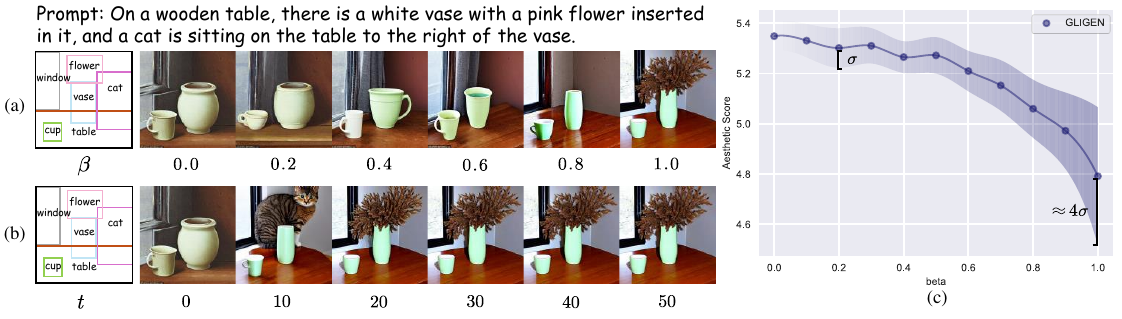}}
\vskip -0.1in
\caption{\textbf{Motivations of RealCompo}. \textbf{(a)} and \textbf{(c)} The realism and aesthetic quality of generated images become poor as more layout is incorporated. \textbf{(b)} Even if layout is incorporated only in the early denoising stages, the control of text alone still fails to alleviate the poor realism issue.}
\end{center}
\vskip -0.4in
\end{figure*}

We conducted experiments to analyze why a significant decrease in image realism exists. We analyze the layout injection mechanism in GLIGEN \citep{li2023gligen} by controlling the density of layout through parameter $\beta$. As shown in Fig. \ref{fig:motivation} (a) and (c), our experiments indicate that the density of layout directly influences the realism of generated images. As the control of layout gradually increases, the generated images become less aesthetic and more unstable. This demonstrates that layout and text, as different control conditions, guide the model towards different generation directions, with the former emphasizing compositionality and the latter emphasizing realism. To alleviate this issue, some models \citep{lian2023llm, li2023gligen} leverage the early-stage localization capability of diffusion models \citep{yu2023freedom, tumanyan2023plug} and incorporate layouts only during the initial denoising phase. In the later denoising stage, only use text  to balance image realism. However, we found this approach yielded minimal effectiveness. We assumed $\beta = 1$ in the first $t$ denoising steps and $\beta = 0$ in the subsequent denoising steps. As shown in Fig. \ref{fig:motivation} (b), the object's position is already determined around $20$ steps. However, it is common that the generated images exhibit almost no difference between $t=20$ and $t=50$. This suggests that even when the injection of layout is stopped in the later denoising stages, the control of text alone still fails to alleviate the poor realism issue. The trade-off between realism and compositionality in T2I and L2I models is challenging yet necessary.

To this end, we introduce a general \textit{training-free} and \textit{transferred-friendly} text-to-image generation framework \textbf{\textit{RealCompo}}, which utilizes a novel \textit{balancer} to achieve dynamic equilibrium between realism and compositionality in generated images. We first utilize LLMs to generate scene layouts from text prompt through in-context learning \citep{min2022rethinking}. Then we propose an innovative \textit{balancer} to dynamically compose pre-trained fidelity-aware (T2I, stylized T2I) and spatial-aware (e.g., layout, keypoint, segmentation map) image diffusion models. This balancer automatically adjusts the coefficient of the predicted noise for each model by analyzing their cross-attention maps during the denoising stage. By combining the respective strengths of the two models, it achieves a trade-off between realism and compositionality. Finally, we extend RealCompo to various spatial-aware conditions through a general compositional denoising process. Moreover, by changing the T2I model to a stylized T2I model, Realcompo can seamlessly achieve compositional generation specified with a particular style. These dramatically demonstrate the great generalization ability of RealCompo. Although there exist methods \citep{xue2023raphael, balaji2022ediffi} for composing multiple diffusion models, their application lacks flexibility because they require additional training and cannot be generalized to other conditionss and models. Our method effectively composes two models in a training-free manner, allowing for a seamless transition between various models.

To the best of our knowledge, RealCompo effectively achieves a trade-off between realism and compositionality in text-to-image generation. Choosing one (stylized) T2I model and one spatial-aware (e.g., layout, keypoint, segmentation map) image diffusion model, RealCompo automatically balances their fidelity and spatial-awareness to realize a collaborative generation. We believe RealCompo opens up a new research perspective in controllable and compositional image generation.

Our main contributions are summarized as the following:
\begin{itemize}
    \item We introduce a new \textit{training-free} and \textit{transferred-friendly} text-to-image generation framework RealCompo, which enhances compositional text-to-image generation by balancing the realism and compositionality of generated images. 
    \item We design a novel \textit{balancer} to dynamically combine the predict noise from T2I model and spatial-aware (e.g., layout, keypoint, segmentation map) image diffusion model.
    \item RealCompo has strong flexibility, can be generalized to balance various (stylized) T2I models and spatial-aware image diffusion models and can achieve high-quality compositional stylized generation. It provides a fresh perspective for compositional image generation.
    \item Extensive qualitative and quantitative comparisons with previous outstanding methods demonstrate that RealCompo has significantly improved the performance in generating multiple objects and complex relationships.
\end{itemize}

\section{Related Work}
\paragraph{Text-to-Image Generation} In recent years, the field of text-to-image generation has made remarkable progress \citep{sun2023dreamsync, xu2023ufogen, podell2023sdxl, hao2024optimizing, du2024stable, zhang2024real, yang2023diffusion}, largely attributed to breakthroughs in diffusion models. By training on large-scale image-text paired datasets, T2I models such as Stable Diffusion (SD) \citep{rombach2022high}, DALL-E 2/3 \citep{ramesh2022hierarchical, betker2023improving}, MDM \citep{gu2023matryoshka}, and Pixart-$\alpha$ \citep{chen2023pixart}, have demonstrated  remarkable generative capabilities. However, there is still significant room for improvement in compositional generation when text prompts include multiple objects and complex relationships \citep{wu2023self}. Many studies have attempted to address this issue through controllable generation \citep{zhang2023adding} by providing additional conditions such as segmentation map \citep{huang2023collaborative}, scene graph \citep{yang2022diffusion}, layout \citep{zheng2023layoutdiffusion}, etc., to constrain the model's generative direction to ensure the accuracy of the number and position of objects in the generated images. However, due to the constraints of the additional conditions, image realism may decrease \citep{li2023gligen}. Furthermore, several works \citep{qu2023layoutllm, chen2023reason, ye2023progressive,yang2024mastering, lu2024llmscore} have attempted to bridge the language understanding gap in models by pre-processing prompts with Large Language Models (LLMs) \citep{achiam2023gpt,touvron2023llama}. It is challenging for T2I models to achieve trade-off between realism and compositionality \citep{yang2024mastering} of generated images.

\paragraph{Compositional Text-to-Image Generation} 
Recently, numerous methods have been introduced to improve compositional text-to-image generation \citep{wang2024instancediffusion, zhou2024migc, yeh2024gen4gen, wen2023improving, li2024mulan, liu2023cones}. These methods enhance diffusion models in attribute binding, object relationship, numeracy, and complex prompts. Recent studies can generally be divided into two types \citep{wang2023compositional}: one primarily uses cross-attention maps for compositional generation \citep{meral2023conform, kim2023dense, zhao2023local}, while the other provides more conditions (e.g., layout, keypoint, segmentation map) to achieve controllable generation \citep{gani2023llm, zhou2024migc}. The first methods delve into a detailed analysis of cross-attention maps, particularly emphasizing their correspondence with the text prompt. Attend-and-Excite \citep{chefer2023attend} dynamically intervenes in the generation process to improve the model's generation results in terms of attribute binding (such as color). Most of the second methods offer layout as a constraint, enabling the model to generate images that meet this condition. This approach directly defines the area where objects are located, making it more straightforward and observable compared to the first type of methods \citep{li2023gligen}. LMD \citep{lian2023llm} provides an additional layout as input with LLMs. Afterward, a controller is designed to predict the masked latent for each object's bounding box and combine them in the denoising process. However, these algorithms are unsatisfactory in the realism of generated images. A recent powerful framework RPG \citep{yang2024mastering} utilizes Multimodal LLMs to decompose complex generation tasks into simpler subtasks to obtain satisfactory realism and compositionality of generated images. Orthogonal to this work, we achieve dynamic equilibrium between realism and compositionality by combining T2I and spatial-aware image diffusion models.

\section{Method}
In this section, we introduce our method, RealCompo, which designs a novel balancer to achieve dynamic equilibrium between realism and compositionality of generated images. We initially focus on the layout-to-image models. In \cref{sec1}, we analyze the necessity of incorporating influence for the predictive noise of each model and provide a method for calculating coefficients. In \cref{sec2}, we provide a detailed explanation of the update rules employed by the balancer, which utilizes a training-free approach to update coefficients dynamically. In \cref{sec3}, we provide a universal formula and denoising procedure that enable the balance of T2I models with any spatial-aware image diffusion model, such as keypoint or segmentation-to-image models based on ControlNet \citep{zhang2023adding}. We also extend RealCompo to stylized compositional generation by stylized T2I models.

\subsection{Combination of Fidelity and Spatial-Awareness}
\label{sec1} 
\begin{figure*}[t!]
\vskip 0.2in
\begin{center}
\centerline{\includegraphics[width=1\textwidth]{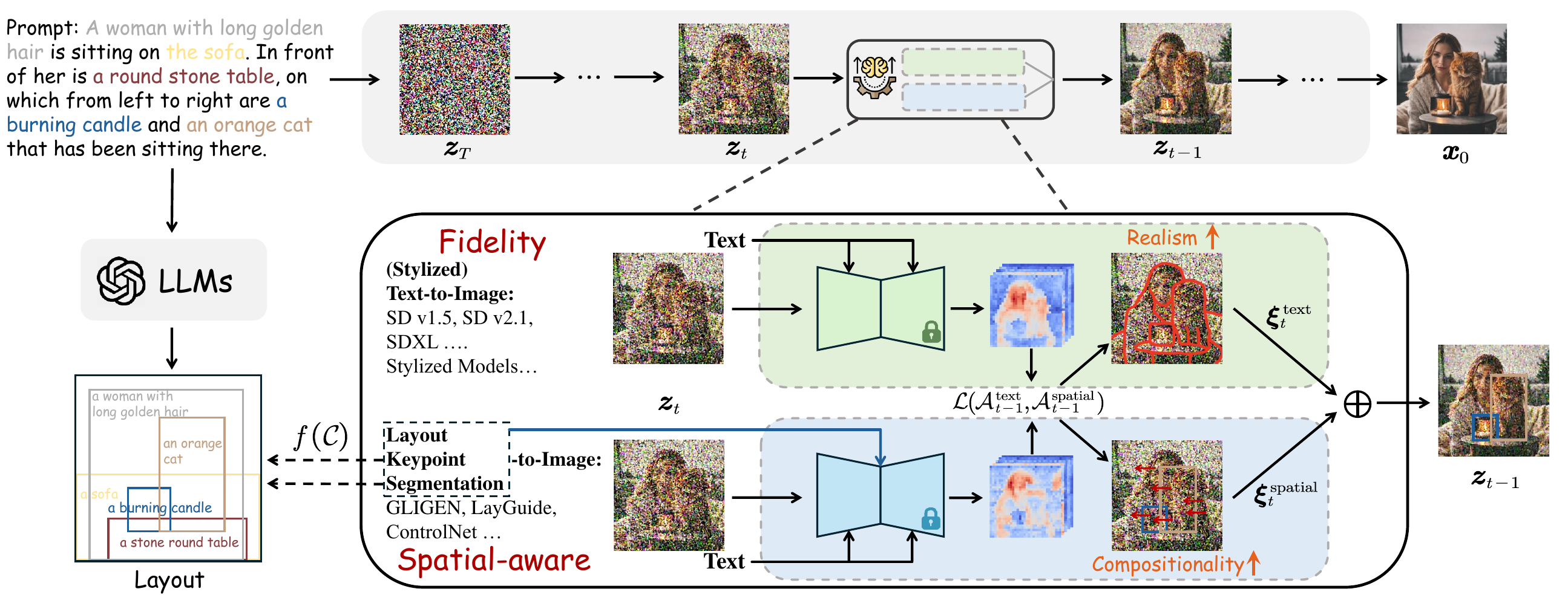}}
\vskip -0.05in
\caption{An overview of RealCompo framework for text-to-image generation. We first use LLMs or transfer function to obtain the corresponding layout. Next, the balancer dynamically updates the influence of two models, which enhances realism by focusing on contours and colors in the fidelity branch, and improves compositionality by manipulating object positions in the spatial-aware branch.}
\label{method}
\end{center}
\vskip -0.3in
\end{figure*}

\paragraph{LLM-based Layout Generation.}Since spatial-aware conditions are similar essentially, we first choose layout as the representative of spatial-aware condition for introduction. As shown in Fig. \ref{method}, we leverage the powerful in-context learning \cite{wu2023autogen} capability of Large Language Models (LLMs) to analyze the input text prompt and generate an accurate layout to achieve "pre-binding" between objects and attributes. The layout is then used as input for the L2I model. In this paper, we choose GPT-4 for layout generation. Please refer to \cref{LLM-based} for detailed explanation.

\paragraph{Combination of Two Types of Noise.}In diffusion models, the model's predicted noise $\boldsymbol{\epsilon}_t$ directly affects the direction of the generated images. In T2I models, $\boldsymbol{\epsilon}^{\text{text}}_t$ exhibits more directive toward realism \citep{rombach2022high}, whereas in L2I models, $\boldsymbol{\epsilon}^{\text{layout}}_t$ demonstrates more directive toward compositionality \cite{li2023gligen}. To achieve the trade-off between realism and compositionality, a feasible but untapped solution is to compose the predicted noise of two models. However, the predicted noise from different models has its own generative direction, contributing differently to the generated results at different timesteps and positions. Based on this, we design a novel balancer that achieves dynamic equilibrium between the two models' strengths at every position $i$ in the noise for timestep $t$. This is achieved by analyzing the influence of each model's predicted noise. Specifically, we first set the same coefficient for the predicted noise of each model to represent their influence before the first denoising step:
\begin{equation}
    \boldsymbol{Coe}^{\text{text}}_T=\boldsymbol{Coe}^{\text{layout}}_T\sim \mathcal{N}( \mathbf{0},\mathbf{I})
\end{equation}
In order to regularize the influence of each model, we perform a softmax operation on the coefficients to get the final coefficients:
\begin{equation}
\label{eq8}
    \boldsymbol\xi^c_t=\frac{\exp({\boldsymbol{Coe}^{c}_t})}{\exp({\boldsymbol{Coe}^{\text{text}}_t})+\exp({\boldsymbol{Coe}^{\text{layout}}_t})}
\end{equation}
where $c\in\{\text{text}, \text{layout}\}$.

The balanced noise can be derived according to the coefficient of each model:
\begin{equation}
\label{eq9}
    \boldsymbol\epsilon_t = \boldsymbol\xi_t^{\text{text}}\odot\boldsymbol\epsilon_t^{\text{text}}+\boldsymbol\xi_t^{\text{layout}}\odot\boldsymbol\epsilon_t^{\text{layout}}
\end{equation}
where $\odot$ denotes pixel-wise multiplication.

Once the predicted noise $\boldsymbol\epsilon_t^{c}$ and the coefficient $\boldsymbol{Coe}^{c}_t$ of each model are provided, the balanced noise can be derived from Eq. \ref{eq8} and Eq. \ref{eq9}. At timestep $t$, the balancer dynamically updates coefficients as described in \cref{sec2}.

\subsection{Influence Estimation with Dynamic Balancer}
\label{sec2}
The alignment between the generated images and the input prompts is largely influenced by model's cross-attention maps, which encapsulate a wealth of matching information between visual and textual elements, such as location and shape. Specifically, given the intermediate feature $\varphi(\boldsymbol{z}_t)$ and the text embeddings $\tau_\theta(y)$, cross-attention maps can be derived in the following manner:
\begin{equation}
    \mathcal{A}^c =\mathrm{Softmax} \left( \frac{Q^c(K^{c})^T}{\sqrt{d^c_k}} \right), c\in\{\text{text}, \text{layout}\}
\end{equation}
\begin{equation}
    Q=W_Q\cdot \varphi \left( \boldsymbol{z}_t \right) , \ K=W_K\cdot \tau _{\theta}(y)
\end{equation}
where $Q$ and $K$ are respectively the dot product results of the intermediate feature $\varphi(\boldsymbol{z}_t)$, text embeddings $\tau_\theta(y)$, and two learnable matrices $W_Q$ and $W_K$. $\mathcal{A}_{ij}$ defines the weight of the value of the $j$-th token on the $i$-th pixel.  Here, $j\in\{1,2,\dots,N(\tau_\theta(y))\}$, and $N(\tau_\theta(y))$ denotes the number of tokens in $\tau_\theta(y)$. The dimension of $K$ is represented by $d_k$.

\paragraph{Update Rule of Dynamic Balancer.}We designed a novel balancer that dynamically balances two models according to their cross-attention maps at timestep $t$. Specifically, we represent layout as $\mathcal{B}=\{b_1,b_2,\dots,b_v\}$, which is composed of $v$ bounding boxes $b$. Each bounding box $b$ corresponds to a binary mask $\mathcal{M}_b$, where the value inside the box is $1$ and the value outside the box is $0$. Given the predicted noise $\boldsymbol\epsilon_t^{c}$ and the coefficient $\boldsymbol{Coe}^{c}_t$ of each model, the balanced noise $\boldsymbol{\epsilon}_t$ and denoised latent $\boldsymbol{z}_{t-1}$ can be derived from Eq. \ref{eq9} and Eq. \ref{eq3}. By feeding $\boldsymbol{z}_{t-1}$ into two models, we obtain the cross-attention maps $\mathcal{A}_{t-1}^{c}$ output by the two models at timestep $t-1$, which indicates the denoising quality feedback after the noise $\boldsymbol{\epsilon}^c_t$ of the model at time $t$ is weighted by $\boldsymbol{\xi}^c_t$. Based on $\mathcal{A}_{t-1}^{c}$, we define the loss function as follows:
% \begin{equation}
% \label{eq10}
%     \mathcal{L}(\mathcal{A}_{t-1}^{\text{text}}, \mathcal{A}_{t-1}^{\text{layout}})=\!\sum_{c}{\sum_{b}{\left( 1\!-\!\frac{\sum_{i}{\mathcal{A}_{(ij_b,t-1)}^{c}\odot \mathcal{M}_b}}{\sum_{i}{\mathcal{A}_{(ij_b,t-1)}^{c}}} \right)}}
% \end{equation}
\begin{equation}
\begin{aligned}
\label{eq10}
     \mathcal{L}(\mathcal{A}_{t-1}^{\text{text}},\mathcal{A}_{t-1}^{\text{layout}})={\sum_{c}\sum_{b}{\left( 1\!-\!\frac{\sum_{i}{\mathcal{A}_{(ij_b,t-1)}^{c}\odot \mathcal{M}_b}}{\sum_{i}{\mathcal{A}_{(ij_b,t-1)}^{c}}} \right)}} 
    \end{aligned}
\end{equation}
where $c\in\{\text{text}, \text{layout}\}$, $j_b$ denotes the token corresponding to the object in bounding box $b$. Since two models are controlled by different conditions, averaging the predicted noise equally will lead to instability in the generated images. This is because the T2I model breaks the layout constraints of the L2I model, reducing the compositionality of the generated images, as we have demonstrated in experimrnts in Fig. \ref{ablation}. Therefore, we designed this loss function to measure the alignment between the cross-attention maps and layout for each model. A smaller loss indicates better compositionality. The following rule is used to update $\boldsymbol{Coe}_t^c$:
\begin{equation}
\label{eq11}
    \boldsymbol{Coe}^{c}_t=\boldsymbol{Coe}^{c}_t-\rho _t\nabla _{\boldsymbol{Coe}^{c}_t}\mathcal{L}(\mathcal{A}_{t-1}^{\text{text}}, \mathcal{A}_{t-1}^{\text{layout}})
\end{equation}
where $\rho_t$ is the updating rate. This update rule continuously strengthens the constraints on both models by assessing the positional alignment of the layout within the cross-attention maps, ensuring the maintenance of the localization capability of L2I model while injecting fidelity information of T2I model. It is worth noting that previous methods \cite{chefer2023attend, xie2023boxdiff,lian2023llm} for parameter updates based on function gradients were primarily using energy functions to update latent $\boldsymbol{z}_t$. We are the first to update the influence of predicted noise based on the gradient of the loss function, which is a novel and stable method well-suited to our task. The complete denoising process is detailed in \cref{inference details}.

\subsection{Extend RealCompo to any Spatial-Aware Conditions in a General Form}
\label{sec3}
\begin{wrapfigure}{r}{0.5\textwidth}
\vspace{-4mm}
    \centering
    \includegraphics[width=.5\textwidth,trim={0 0 0 0},clip]{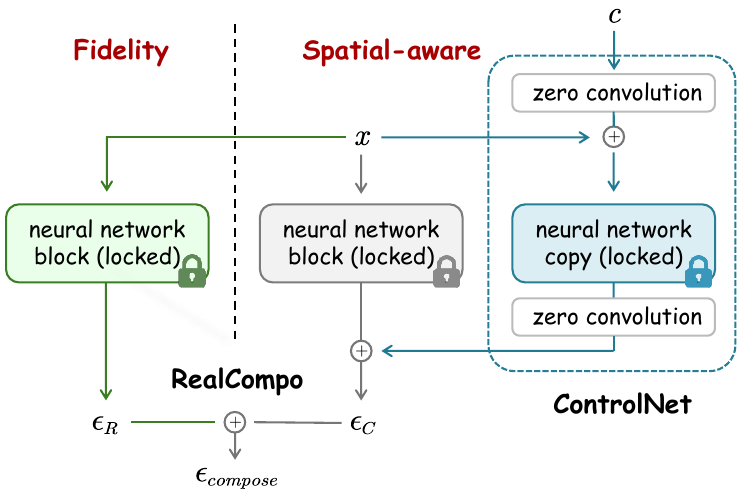}
    \vspace{-3mm}
    \caption{RealCompo constructed on ControlNet.}
    \label{fig:control}
    \vspace{-3mm}
\end{wrapfigure}

Other spatial-aware text-to-image diffusion models are essentially similar to L2I models. Keypoint-to-image (K2I) models generate specified actions or poses within each group of keypoints region, and segmentation-to-image (S2I) models fill indicated objects within each segmented region. The concept of "region" is always present, which transforms T2I generation from a macro perspective to utilizing region-based control for T2I generation from a micro perspective. This concept is also the core of enhancing image compositionality. Compared with layout-based T2I generation, the only difference is that keypoints and segmentation maps have stronger control over the model based on regions, requiring that the pose is maintained and the object is correct and unique.

\begin{figure*}[t!]
\vskip 0.0in
\begin{center}
\centerline{\includegraphics[width=1\textwidth]{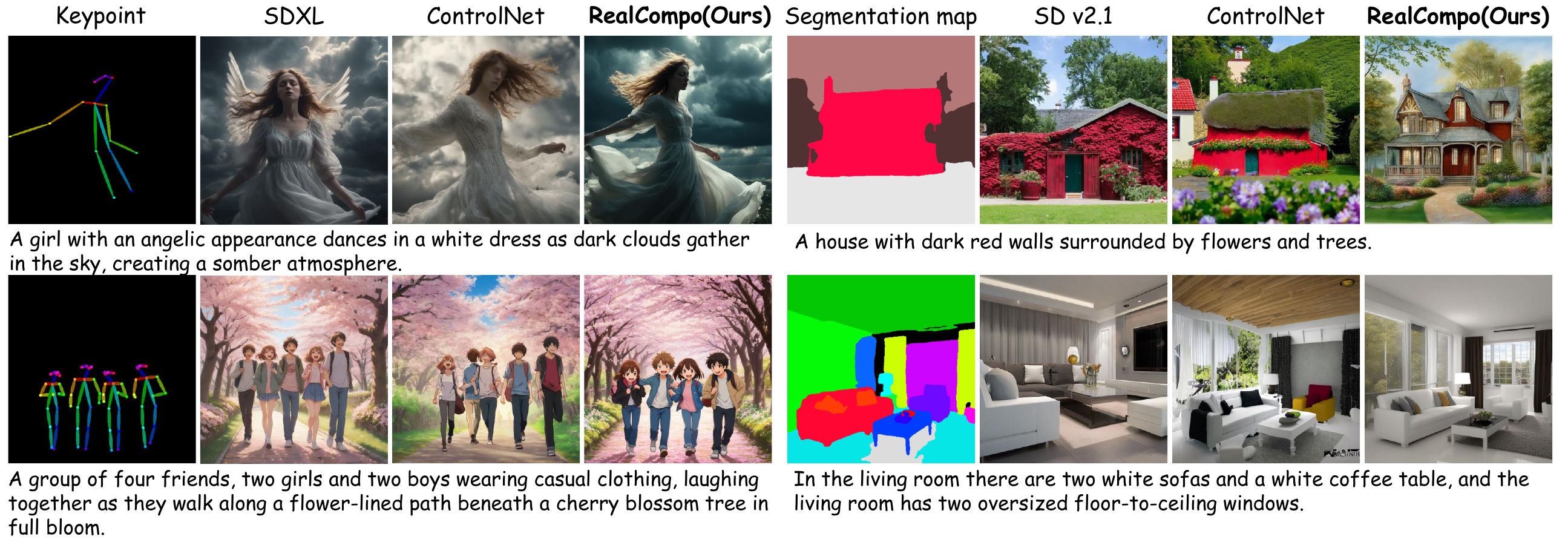}}
\vskip -0.1in
\caption{Extend RealCompo to keypoint- and segmentation-based image generation.}
\label{key_and_seg}
\end{center}
\vskip -0.2in
\end{figure*}

\paragraph{General Form for Extension to Other Spatial-Aware Conditions}We rethink Eq. \ref{eq10}, which is RealCompo's core approach in combining T2I and L2I models, where the only layout-related variable is the binary masks $\mathcal{M}$. Considering that spatial-aware controllable T2I generation inherently focus on the concept of "region control", we introduce a transfer function:
\begin{equation} \label{transfer}
    \mathcal{M} = f(\mathcal{C})
\end{equation}
where $\mathcal{C}$ represents other spatial-aware conditions such as keypoint and segmentation map. $f(\cdot)$ represents the calculation of the minimum and maximum values of the horizontal and vertical coordinates occupied by each set of keypoints or a segmentation block within the entire image coordinate system, which can be transformed into a layout and a binary mask $\mathcal{M}$. Therefore, for any T2I models with spatial-aware control, the general loss function of RealCompo is:
\begin{equation}
\label{eq13}
    \mathcal{L}(\mathcal{A}_{t-1}^{\text{text}},\mathcal{A}_{t-1}^{\text{spatial}})={\sum_{c}\sum_{b}{\left( 1\!-\!\frac{\sum_{i}{\mathcal{A}_{(ij_b,t-1)}^{c}\odot f_b(\mathcal{C})}}{\sum_{i}{\mathcal{A}_{(ij_b,t-1)}^{c}}} \right)}} 
\end{equation}
where $c\in\{\text{text}, \text{spatial}\}$. Similarly, $\boldsymbol{Coe}_t^c$ is dynamically updated using Eq. \ref{eq11}. ControlNet \citep{zhang2023adding} enables controllable T2I generation based on various spatial-aware conditions. In this work, the spatial-aware branches besides layout are all based on ControlNet, which is illustrated in Fig. \ref{fig:control}. The generated images of keypoint- and segmentation-based RealCompo are shown in Fig. \ref{key_and_seg}.

\paragraph{Extend RealCompo to Stylized Image Generation}As an essential indicator of fidelity, image style \citep{wang2024instantstyle, ye2023ip} guides us to expand the application potential of RealCompo. Since RealCompo mainly leverages T2I models to enhance and guide the realism and aesthetic quality of generated images. By replacing the T2I model with various stylized T2I models and combining it with a spatial-aware image diffusion model, we can achieve outstanding compositional generation under this style. The experiments are shown in Fig \ref{style}.

\section{Experiments}
\subsection{Experimental Setup}

\paragraph{Implementation Details}
Our RealCompo is a generic, scalable framework that can achieve the complementary advantages of the model with any chosen (stylized) T2I models and spatial-aware image diffusion models. We selected GPT-4 \cite{achiam2023gpt} as the layout generator in our experiments, the detailed rules are described in \cref{LLM-based}. For layout-based RealCompo, we chose SD v1.5 \cite{rombach2022high} and GLIGEN \cite{li2023gligen} as the backbone. For keypoint-based RealCompo, we chose SDXL \cite{betker2023improving} and ControlNet \cite{zhang2023adding} as the backbone. For segmentation-based RealCompo, we chose SD v2.1 \cite{rombach2022high} and ControlNet \cite{zhang2023adding} as the backbone. For style-based RealCompo, we chose two stylized T2I models: Coloring Page Diffusion and CuteYukiMix as the backbone, and chose GLIGEN \cite{li2023gligen} as the backbone of L2I model. All of our experiments are conducted under 1 NVIDIA 80G-A100 GPU.

\paragraph{Baselines and Benchmark} 

To evaluate compositionality, we compare our RealCompo with the outstanding T2I and L2I models on T2I-CompBench \citep{huang2023t2i}. This benchmark test models across aspects of attribute binding, object relationship, numeracy and complexity. To evaluate realism, we randomly select 3K text prompts from the COCO validation set , we utilize ViT-B-32 \citep{dosovitskiy2020image} to calculate the CLIP score and LAION aesthetic predictor to calculate aesthetic score, reflecting the degree of match between generated images and prompts as well as the aesthetic quality, respectively. In addition to objective evaluations, we conducted a user study to evaluate RealCompo and stylized RealCompo in terms of realism, compositionality, and comprehensive evaluation.

\label{experiment setup}

\begin{table*}[t!]
% \vspace{-1.5em}
\centering
\caption{Evaluation results about compositionality on T2I-CompBench \cite{huang2023t2i}. 
RealCompo consistently demonstrates the best performance regarding attribute binding, object relationships, numeracy and complex compositions.
We denote the best score in \colorbox{pearDark!20}{blue}, and the second-best score in \colorbox{mycolor_green}{green}. The baseline data is quoted from PixArt-$\alpha$ \citep{chen2023pixart}.} 
\vspace{0.5em}
\label{benchmark:t2icompbench}
\resizebox{1\linewidth}{!}{ 
\begin{tabular}
{lccccccc}
\toprule
\multicolumn{1}{c}
{\multirow{2}{*}{\bf Model}} & \multicolumn{3}{c}{\bf Attribute Binding } & \multicolumn{2}{c}{\bf Object Relationship} & \multirow{2}{*}{\bf Numeracy$\uparrow$}& \multirow{2}{*}{\bf Complex$\uparrow$}
\\
\cmidrule(lr){2-4}\cmidrule(lr){5-6}

&
{\bf Color $\uparrow$ } &
{\bf Shape$\uparrow$} &
{\bf Texture$\uparrow$} &
{\bf Spatial$\uparrow$} &
{\bf Non-Spatial$\uparrow$} &
\\
\midrule
Stable Diffusion v1.4 \cite{rombach2022high}   & 0.3765 & 0.3576 & 0.4156 & 0.1246 & 0.3079 & 0.4461 & 0.3080  \\
Stable Diffusion v2 \cite{rombach2022high}  & 0.5065 & 0.4221 & 0.4922 & 0.1342 & 0.3096 & 0.4579 & 0.3386  \\
Structured Diffusion \cite{feng2022training} & 0.4990 & 0.4218 & 0.4900 & 0.1386 &  0.3111 & 0.4550 & 0.3355  \\
Attn-Exct v2 \cite{chefer2023attend} &  0.6400 & 0.4517 & 0.5963 & 0.1455 & 0.3109 & 0.4767& 0.3401  \\
DALL-E 2 \cite{ramesh2022hierarchical}  & 0.5750 & 0.5464 & 0.6374 & 0.1283 & 0.3043 & 0.4873 & 0.3696  \\
Stable Diffusion XL \cite{betker2023improving} & 0.6369 & 0.5408 & 0.5637 & 0.2032 & 0.3110 & 0.4988 & 0.4091  \\
PixArt-$\alpha$ \cite{chen2023pixart} & \cellcolor{mycolor_green}{0.6886} & \cellcolor{mycolor_green}{0.5582} & \cellcolor{mycolor_green}{0.7044} & {0.2082} & \cellcolor{mycolor_green}{0.3179} & 0.5058 & \cellcolor{mycolor_green}{0.4117}  \\
\midrule
GLIGEN\cite{li2023gligen} & 0.4288  & 0.3998 & 0.3904 & \cellcolor{mycolor_green}{0.2632} & 0.3036 & 0.4970 & 0.3420 \\
LMD+\cite{lian2023llm} & 0.4814  & 0.4865 & 0.5699 & 0.2537& 0.2828 &  \cellcolor{mycolor_green}{0.5762} & 0.3323 \\
\midrule
\textbf{RealCompo (Ours)} & \cellcolor{pearDark!20}{0.7741} & \cellcolor{pearDark!20}{0.6032} &  \cellcolor{pearDark!20}{0.7427} & \cellcolor{pearDark!20}{0.3173} & \cellcolor{pearDark!20}{0.3294}  & \cellcolor{pearDark!20}{0.6592} & \cellcolor{pearDark!20}{0.4657} \\
\bottomrule
\end{tabular}
}
\vspace{-1em}
\end{table*}

\begin{figure*}[b]
\vskip -0.2in
\begin{center}
\centerline{\includegraphics[width=1\textwidth]{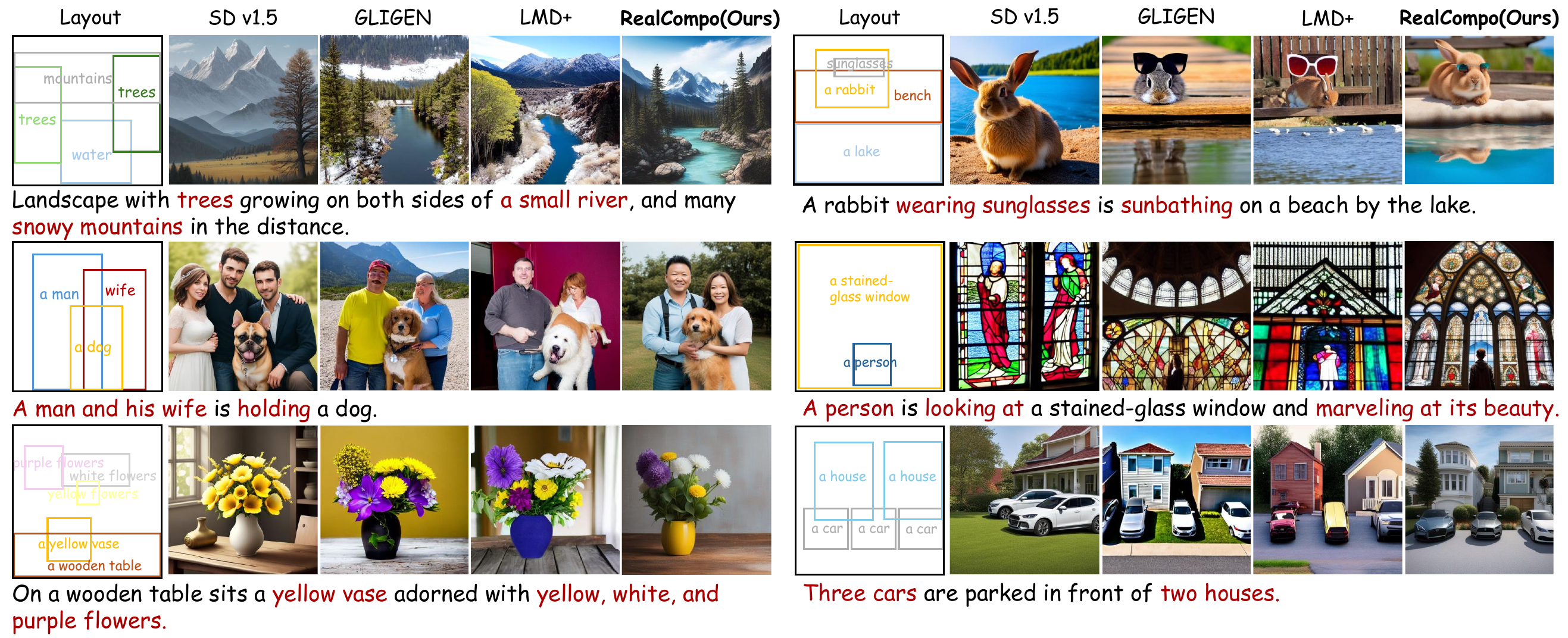}}
\vskip -0.1in
\caption{Qualitative comparison between our RealCompo and the outstanding text-to-image model Stable Diffusion v1.5 \cite{rombach2022high}, as well as the layout-to-image models, GLIGEN \cite{li2023gligen} and LMD+ \cite{lian2023llm}. Colored text denotes the advantages of RealCompo in generated images.}
\label{experiment}
\end{center}
\vskip -0.3in
\end{figure*}

\subsection{Main Results}

\paragraph{Results of Compositionality: T2I-CompBench}
We conducted tests on T2I-CompBench \citep{huang2023t2i} to evaluate the compositionality of RealCompo compared to the outstanding T2I and L2I models. As demonstrated in Table \ref{benchmark:t2icompbench}, RealCompo achieved state-of-the-art performance on all seven evaluation tasks. It is clear that RealCompo and L2I models GLIGEN \citep{li2023gligen} and LMD+ \citep{lian2023llm} show significant improvements in spatial-aware tasks such as spatial and numeracy. These improvements are largely attributed to the guidance provided by the additional conditions, which greatly enhances the model's compositional performance. RealCompo employs a balancer for better control over positioning, boosting its advantages in these aspects. However, the L2I models exhibit a noticeable decline in performance on tasks like texture and non-spatial. This decline is due to the injection of layout embeddings, which dilute the density of text embeddings, leading to suboptimal semantic understanding by the model. By composing additional T2I models, RealCompo provides sufficient textual information during the denoising process and achieves outstanding results in tasks that reflect realism, such as texture, non-spatial and complex tasks. As shown in Fig. \ref{experiment}, compared with the current outstanding L2I models GLIGEN and LMD+, RealCompo achieves a high level of realism while keeping the attributes of the objects matched and the number of positions generated correctly.

\begin{wraptable}{r}{0.58\textwidth}  % 'r' for right, 0.5\textwidth for half the width of the text
    \centering
    \scriptsize
    \caption{Evaluation results on image realism.}\vspace{0.5em}
    \tabcolsep=0.15cm
    \begin{tabular}{lcc}
        \toprule
        \textbf{Model}  & \textbf{CLIP Score}$\uparrow$ & \textbf{Aesthetic Score}$\uparrow$\\
        \midrule
        Stable Diffusion v1.4 \cite{rombach2022high}  & 0.307  & 5.326  \\
        TokenCompose v2.1 \cite{wang2023tokencompose} &  \cellcolor{mycolor_green}{0.323} & 5.067 \\
        Stable Diffusion v2.1 \cite{rombach2022high}  & 0.321  & 5.458 \\
        Stable Diffusion XL \cite{betker2023improving}  & 0.322 & \cellcolor{mycolor_green}{5.531} \\
        \midrule
        Layout Guidance\citep{chen2024training} & 0.294 & 4.947  \\
        GLIGEN\cite{li2023gligen} & 0.301  & 4.892   \\
        LMD+\citep{lian2023llm} &0.298 &  4.964\\
        \midrule
        \textbf{RealCompo (Ours)}  &  \cellcolor{pearDark!20}{0.334} &  \cellcolor{pearDark!20}{5.742}  \\
        \bottomrule
    \end{tabular}
    \label{realism}
    \vspace{-1em}
\end{wraptable}

\begin{figure*}[t!]
\vskip 0.0in
\begin{center}
\centerline{\includegraphics[width=1\textwidth]{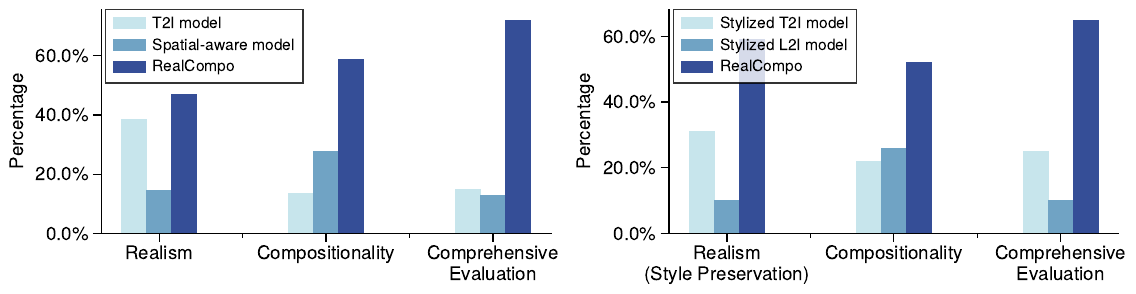}}
\vskip -0.1in
\caption{Results of user study.}
\label{user_study}
\end{center}
\vskip -0.3in
\end{figure*}

\paragraph{Results of Realism: Quantitative Comparison and User Study}As shown in Table \ref{realism}, our model significantly outperforms existing outstanding T2I and L2I models in both CLIP score and aesthetic score. We attribute this to the dynamic balancer, which enhances image realism and aesthetic quality while maintaining high compositionality. In addition to objective evaluations, we designed a user study to subjectively assess the practical performance of various methods. We randomly selected 15 prompts, including 5 for stylization experiments. Comparative tests were conducted using T2I models, spatial-aware image diffusion models, and RealCompo. We invited 39 users from diverse backgrounds to vote on image realism, image compositionality, and comprehensive evaluation, resulting in a total of 1755 votes. As illustrated in Fig. \ref{user_study}, RealCompo received widespread user approval in terms of realism and compositionality.

\begin{figure*}[b]
\vskip -0.2in
\begin{center}
\centerline{\includegraphics[width=0.92\textwidth]{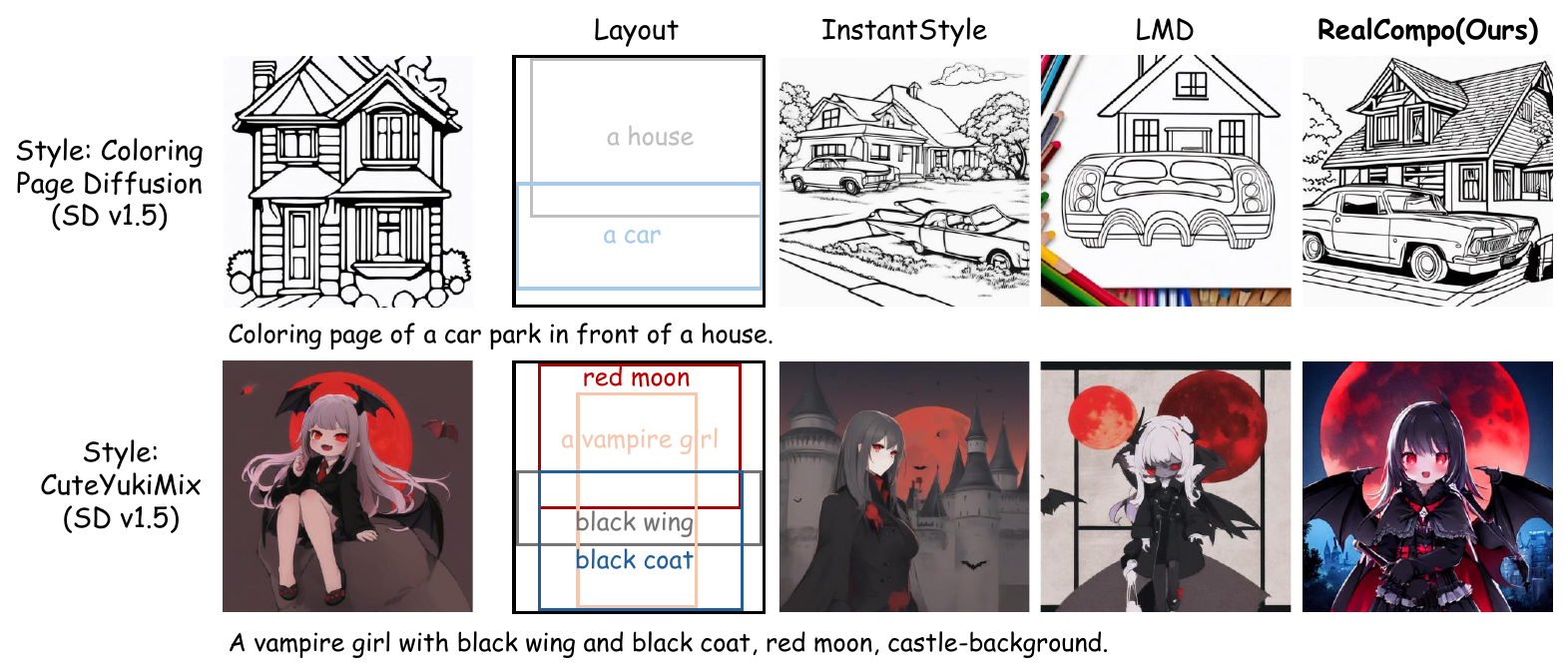}}
\vskip -0.05in
\caption{Extend RealCompo to stylized compositional generation.}
\label{style}
\end{center}
\vskip -0.3in
\end{figure*}

\paragraph{Results of Extend Applications: More Spatial-Aware Conditions}
We extend RealCompo to more spatial-aware controlled image generation. As shown in Fig. \ref{key_and_seg}, keypoint- and segmentation-based RealCompo achieves outstanding performance in both realism and compositionality. This promising result reveals that as spatial-aware conditions, layout, keypoint, and segmentation map are fundamentally similar, RealCompo focuses on these similarities and achieves a general generative paradigm for compositional generation.

\paragraph{Results of Extend Applications: Stylized Generation}Image style is an essential indicator of fidelity. We experiment with generalizing RealCompo to various pre-trained stylized T2I models. We selected the Coloring Page Diffusion and Cutyukimix as the foundational stylized models, focusing on the coloring page style and adorable style, respectively. As shown in Fig. \ref{style}, RealCompo perfectly inherits the style of the T2I models and, with the help of L2I model, achieves powerful compositional generation under these styles, which is currently difficult for stylized diffusion models to accomplish. We found it difficult for LMD to strictly maintain the style by simply replacing the backbone with a stylized model, often leading to text leakage \citep{feng2022training}. For example, terms like "crayon" frequently appear in the coloring page style, indicating that the layout control disrupts the style or text control, making it challenging for L2I models to achieve stylized compositional generation. In contrast, by maintaining image realism and style, RealCompo demonstrates strong compositionality while better preserving the style compared to currently outstanding stylized models like InstantStyle \citep{wang2024instantstyle}.

\begin{figure*}[t!]
\vskip 0.0in
\begin{center}
\centerline{\includegraphics[width=0.95\textwidth]{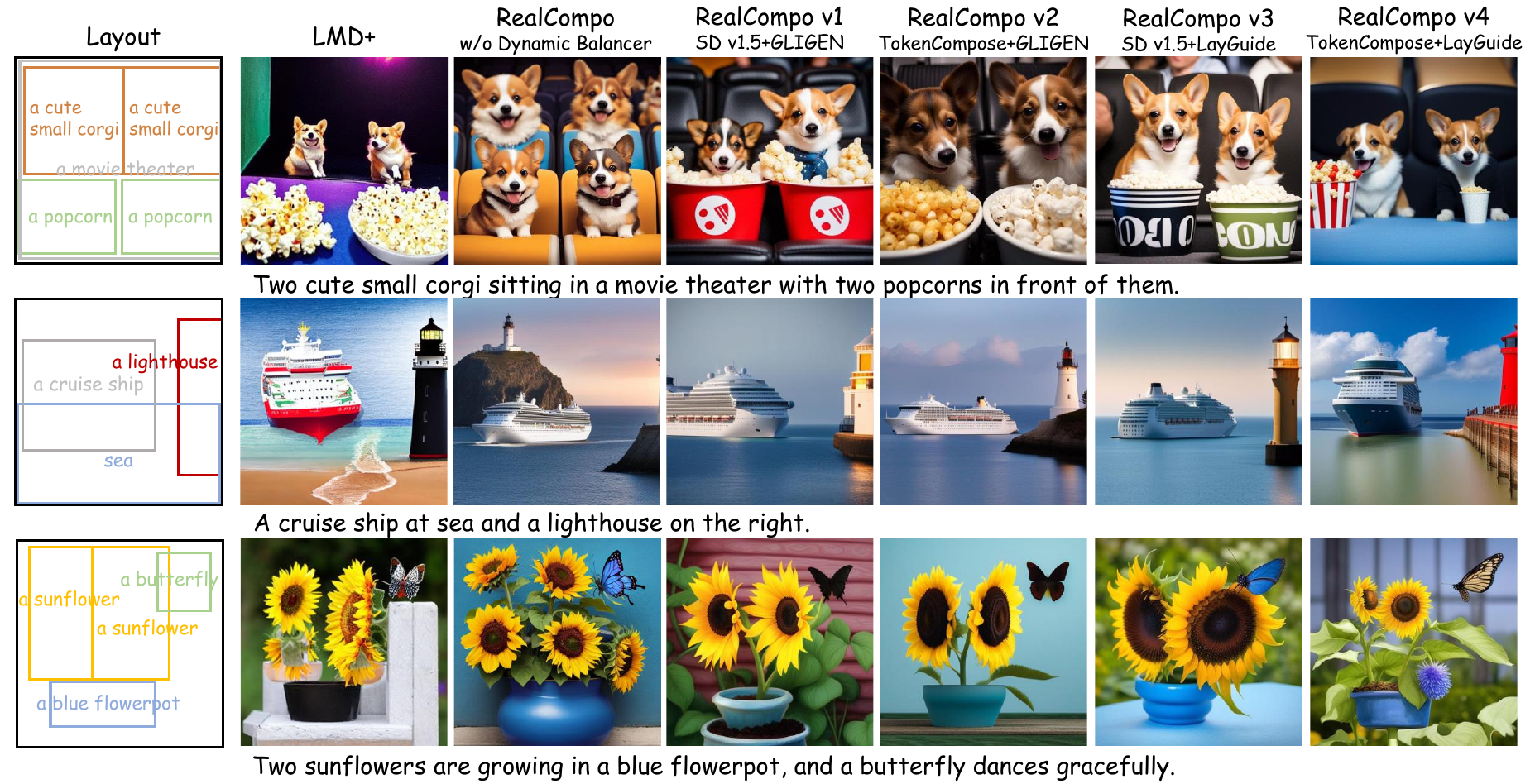}}
\vskip -0.1in
\caption{Ablation study on the significance of the dynamic balancer and qualitative comparison of RealCompo's generalization to different models. We demonstrate that dynamic balancer is important to compositional generation and RealCompo has strong generalization and generality to different models, achieving a remarkable level of both fidelity and precision in aligning with text prompts.}
\label{ablation}
\end{center}
\vskip -0.4in
\end{figure*}

\subsection{Ablation Study}\label{ablation exp}
% \paragraph{Qualitative Comparisons}
\paragraph{Importance of Dynamic Balancer} As shown in Fig. \ref{ablation}, we conducted experiments on the importance of the dynamic balancer. It is clear that without the use of the dynamic balancer, the generated images do not align with the layout. This is because the predicted noise in T2I model is not constrained by the layout, leading to the model generating the object at any position, and the quantity is uncontrollable. Although the image realism is high, the predicted noise of T2I model disrupts the object distribution of the predicted noise of L2I model, leading to poor compositionality of the generated images and uncontrollable in the generation process.

\paragraph{Generalizing to Different Backbones} To explore the generalizability of RealCompo for various models, we choose two T2I models, SD v1.5 \cite{rombach2022high} and TokenCompose \cite{wang2023tokencompose}, and two L2I models, GLIGEN \cite{li2023gligen} and LayGuide (Layout Guidance) \cite{chen2024training}. We combine them two by two, yielding four versions of RealCompo v1-v4. The experimental results are shown in Fig. \ref{ablation}. The four versions of RealCompo all have a high degree of realism in generating images and achieving desirable results regarding instance composition. This is attributed to the dynamic balancer combining the strengths of T2I and L2I models, and it can seamlessly switch between models because it is simple and requires no training. We also found that RealCompo, when using GLIGEN as the L2I model, performs better than when using LayGuide in generating objects that match the layout. For instance, in the images generated by RealCompo v4 in the first and third rows, "popcorns" and "sunflowers" do not fill up the bounding box, which can be attributed to the superior performance of the base model GLIGEN compared to LayGuide. Therefore, when combined with more powerful T2I and L2I models, RealCompo is expected to yield more satisfactory results.

\section{Conclusion}

In this paper, to solve the challenge of complex or compositional text-to-image generation, we propose the SOTA training-free and transferred-friendly framework RealCompo. In RealCompo, we propose a novel balancer that dynamically combines the advantages of various (stylized) T2I and spatial-aware (e.g., layout, keypoint, segmentation map) image diffusion models to achieve the trade-off between realism and compositionality in generated images. In future work, we will continue to improve this framework by using a more powerful backbone and extend it to more realistic applications.

{\small
\bibliographystyle{ieee}
\bibliography{neurips_2024}
}

%%%%%%%%%%%%%%%%%%%%%%%%%%%%%%%%%%%%%%%%%%%%%%%%%%%%%%%%%%%%

\newpage

\appendix
This supplementary material is structured into several sections that provide additional details and analysis related to our work on RealCompo. Specifically, it will cover the following topics:
\begin{itemize}
    \item In \cref{preliminary}, we provide a preliminary about Stable Diffusion.
    \item In \cref{LLM-based}, we provide a detailed pipeline about how to get layout through in-context learning of LLMs.
    \item In \cref{analysis_of_gradient}, we provide a detailed proof of the existence of the gradient in Eq. \ref{eq11}.
    \item In \cref{inference details}, we provide the pseudocode for RealCompo to thoroughly demonstrate its denoising process.
    \item In \cref{gradient analysis}, we conduct a detailed analysis of the gradient changes of the two models in Eq. \ref{eq11} during the denoising process.
    \item In \cref{Limitations}, we analysis the limitations and future work of RealCompo.
    \item In \cref{broader impact}, we analysis the broader impact of RealCompo.
    \item In \cref{more results}, we provide more additional visualized results.
    
\end{itemize}
\section{Preliminary}\label{preliminary}
% \subsection{Text-to-Image Generation: Stable Diffusion}
Diffusion models \citep{ho2020denoising, sohl2015deep} are probabilistic generative models. They can perform multi-step denoising on random noise $\boldsymbol{x}_T\sim\mathcal{N}(\mathbf{0},\mathbf{I})$ to generate clean images through training. Specifically, a gaussian noise $\boldsymbol{\epsilon}$ is gradually added to the clean image $\boldsymbol{x}_0$ in the forward process:
\begin{equation}
    \boldsymbol{x}_t=\sqrt{\bar\alpha_t}\boldsymbol{x}_0+\sqrt{1-\bar\alpha_t}\boldsymbol{\epsilon}
\end{equation}
where $\boldsymbol{\epsilon} \sim \mathcal{N}( \mathbf{0},\mathbf{I}) $ and $\alpha_t$ is the noise schedule.

Training is performed by minimizing the squared error loss:
\begin{equation}
    \min_{\boldsymbol{\theta }} \mathcal{L} =\mathbb{E} _{\boldsymbol{x},\boldsymbol{\epsilon }\sim \mathcal{N} (\mathbf{0},\mathbf{I}),t}\left[ \left\| \boldsymbol{\epsilon }-\boldsymbol{\epsilon_\theta }(\boldsymbol{x}_t,t) \right\| _{2}^{2} \right]
\end{equation}
The parameters of the estimated noise $\boldsymbol{\epsilon_\theta}$ are updated step by step by calculating the loss between the real noise $\boldsymbol{\epsilon}$ and the estimated noise $\boldsymbol{\epsilon_\theta }(\boldsymbol{x}_t,t)$.

The reverse process aims to start from the noise $\boldsymbol{x}_T$, and denoise it according to the predicted noise $\boldsymbol{\epsilon_\theta }(\boldsymbol{x}_t,t)$ at each step. DDIM \cite{song2020denoising} is a deterministic sampler with denoising steps:
\begin{equation}\label{eq3}
\boldsymbol{x}_{t-1}=\  \sqrt{\bar{\alpha}_{t-1}}\left( \frac{\boldsymbol{x}_t-\sqrt{1-\bar{\alpha}_t}\boldsymbol{\epsilon }_{\boldsymbol{\theta }}\left( \boldsymbol{x}_t,t \right)}{\sqrt{\bar{\alpha}_t}} \right) +\sqrt{1-\bar{\alpha}_{t-1}}\boldsymbol{\epsilon }_{\boldsymbol{\theta }}\left( \boldsymbol{x}_t,t \right) 
\end{equation}
% The final result is a clean image $x_0$ through this method of continuous denoising.
Stable Diffusion \cite{rombach2022high} is a significant advancement in this field, which conducts noise addition and removal in the latent space. Specifically, SD uses a pre-trained autoencoder that consists of an encoder $\mathcal{E}$ and a decoder $\mathcal{D}$. Given an image $\boldsymbol{x}$, the encoder $\mathcal{E}$ maps $\boldsymbol{x}$ to the latent space, and the decoder $\mathcal{D}$ can reconstruct this image, i.e., $\boldsymbol{z}=\mathcal{E}(\boldsymbol{x})$, $\tilde{\boldsymbol{x}}=\mathcal{D}(\boldsymbol{z})$. Moreover, Stable Diffusion supports an additional text prompt $y$ for conditional generation. $y$ is transformed into text embeddings $\tau_\theta(y)$ through the pre-trained CLIP \cite{radford2021learning} text encoder. $\boldsymbol{\epsilon_\theta}$ is trained via:
\begin{equation}
    \min_{\boldsymbol{\theta }} \mathcal{L} \!=\!\mathbb{E} _{\boldsymbol{z} \sim\mathcal{E}(\boldsymbol{x}),\boldsymbol{\epsilon }\sim \mathcal{N} (\mathbf{0},\mathbf{I}),t}\left[ \left\| \boldsymbol{\epsilon }\!-\!\boldsymbol{\epsilon_\theta }(\boldsymbol{z}_t,t,\tau_\theta(y)) \right\| _{2}^{2} \right]
\end{equation}
In the inference process, noise $\boldsymbol{z}_T \sim \mathcal{N}\left( \boldsymbol{0},\mathbf{I} \right) $ is sampled from the latent space. By applying Eq. \ref{eq3}, we perform step-by-step denoising to obtain a clean latent $\boldsymbol{z}_0 $. The generative image is then reconstructed through the decoder $\mathcal{D}$.

\newpage
\section{Additional Analysis}

\subsection{LLM-based Layout Generation}
\label{LLM-based}
Large Language Models (LLMs) have witnessed remarkable advancements in recent years \cite{touvron2023llama, kazemi2022lambada}. Due to their robust language comprehension, induction, reasoning, and summarization capabilities, LLMs have made significant strides in the Natural Language Processing (NLP) tasks \cite{foley2023matching, wu2023speechgen}. In the context of multiple-object compositional generation, text-to-image diffusion models exhibit a relatively weaker understanding of language, as reflected in the poor compositionality of the generated images. Consequently, exploring ways to harness the inferential and imaginative capacities of LLMs to facilitate their collaboration with text-to-image diffusion models, thereby producing images that adhere to the prompt, offers substantial research potential.

\begin{figure}[ht]
\vskip 0.0in
\begin{center}
\centerline{\includegraphics[width=.7\textwidth]{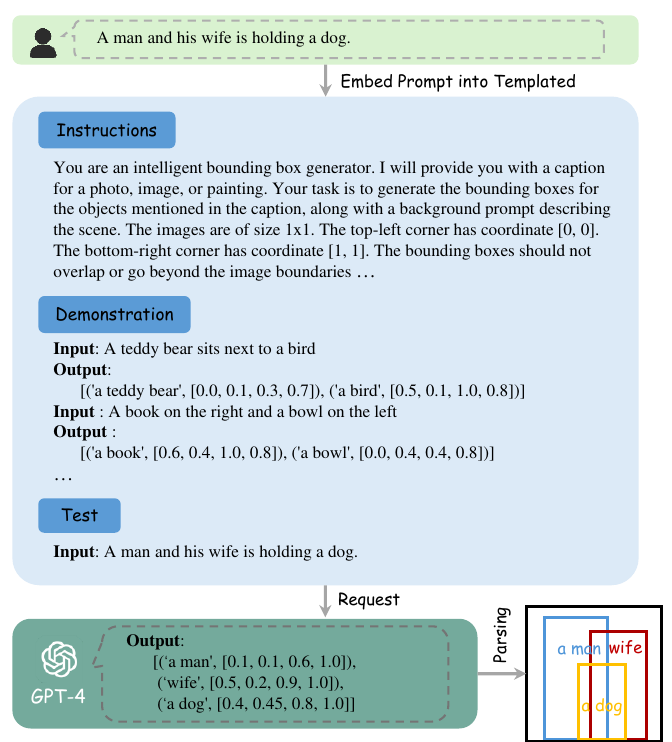}}
\caption{Firstly, the user's input text is embedded into the prompt template. The template is then parsed using GPT-4 with frozen parameters, which yields descriptions of the objects in the prompt as well as their corresponding layout.}
\label{gpt4}
\end{center}
\vskip -0.2in
\end{figure}

In our task, we leverage LLMs to directly infer the layout of all objects based on the user's input prompt through in-context learning (ICL) \cite{li2023unified,si2023measuring}. This layout is used for the layout-to-image model of RealCompo, eliminating the need to manually provide a layout for each prompt and achieve pre-binding of multiple objects and attributes. Specifically, as shown in Fig. \ref{gpt4}, we construct prompt templates, which include descriptions of task rules (instruction), in-context examples (demonstration), and the user's input prompt (test). Through imitation reasoning based on the instruction, LLM generate layout for each object, where each layout represents the coordinates of the top-left and bottom-right corners of a respective box. We selected the highly capable GPT-4 \cite{achiam2023gpt} as layout generator.

\newpage
\subsection{Analysis of the Existence of Gradient in Eq. \ref{eq11}}\label{analysis_of_gradient}
Here we set:
\begin{equation}
\begin{aligned}
    \mathcal{L}(\mathcal{A}_{t-1}^{\text{text}}, \mathcal{A}_{t-1}^{\text{layout}})&=\sum_b\mathcal{L}_b(\mathcal{A}_{t-1}^{\text{text}}, \mathcal{A}_{t-1}^{\text{layout}})\\&={\sum_{b}\left[\left( 1-\frac{\sum_{i}{\mathcal{A}_{(ij_b,t-1)}^{\text{text}}\odot \mathcal{M}_b}}{\sum_{i}{\mathcal{A}_{(ij_b,t-1)}^{\text{text}}}}\right) +{\left( 1-\frac{\sum_{i}{\mathcal{A}_{(ij_b,t-1)}^{\text{layout}}\odot\mathcal{M}_b}}{\sum_{i}{\mathcal{A}_{(ij_b,t-1)}^{\text{layout}}}} \right)}\right]}
\end{aligned}
\end{equation}
If the loss function is given by Eq. \ref{eq10}, the gradient in Eq. \ref{eq11} can be derived as follows:
\begin{equation}
\begin{aligned}
\label{eq14}
&\frac{\partial \mathcal{L} \left( \mathcal{A} _{t-1}^{\text{text}},\mathcal{A} _{t-1}^{\text{layout}} \right)}{\partial \boldsymbol{Coe}_{t}^{c}}\\=&\frac{\partial\sum_b \mathcal{L}_b \left( \mathcal{A} _{t-1}^{\text{text}},\mathcal{A} _{t-1}^{\text{layout}} \right)}{\partial \boldsymbol{Coe}_{t}^{c}}\\=&\sum_b\frac{\partial \mathcal{L}_b \left( \mathcal{A} _{t-1}^{\text{text}},\mathcal{A} _{t-1}^{\text{layout}} \right)}{\partial \boldsymbol{Coe}_{t}^{c}}\\=&\sum_b\left[\frac{\partial \mathcal{L}_b \left( \mathcal{A} _{t-1}^{\text{text}},\mathcal{A} _{t-1}^{\text{layout}} \right)}{\partial \mathcal{A} _{(j_b,t-1)}^{c}}\frac{\partial \mathcal{A} _{(j_b,t-1)}^{c}}{\partial \boldsymbol{z}_{t-1}}\frac{\partial \boldsymbol{z}_{t-1}}{\partial \boldsymbol{\epsilon }_t}\frac{\partial \boldsymbol{\epsilon }_t}{\partial \boldsymbol{\xi }^c_t}\frac{\partial \boldsymbol{\xi }^c_t}{\partial \boldsymbol{Coe}_{t}^{c}}\right] \\=&\sum_b\left[\frac{\partial \mathcal{L}_b \left( \mathcal{A} _{t-1}^{\text{text}},\mathcal{A} _{t-1}^{\text{layout}} \right)}{\partial \mathcal{A} _{(j_b,t-1)}^{c}}\frac{\partial \mathcal{A} _{(j_b,t-1)}^{c}}{\partial \boldsymbol{z}_{t-1}}\frac{\partial \boldsymbol{z}_{t-1}}{\partial \boldsymbol{\epsilon }_t}\frac{\partial \boldsymbol{\epsilon }_t}{\partial \boldsymbol{\xi }^c_t}\frac{\exp \left( \boldsymbol{Coe}_{t}^{\text{text}}+\boldsymbol{Coe}_{t}^{\text{layout}} \right)}{\left( \exp \left( \boldsymbol{Coe}_{t}^{\text{text}} \right) +\exp \left( \boldsymbol{Coe}_{t}^{\text{layout}} \right) \right) ^2}\right] \\=&\sum_b\left[\frac{\partial \mathcal{L}_b \left( \mathcal{A} _{t-1}^{\text{text}},\mathcal{A} _{t-1}^{\text{layout}} \right)}{\partial \mathcal{A} _{(j_b,t-1)}^{c}}\frac{\partial \mathcal{A} _{(j_b,t-1)}^{c}}{\partial \boldsymbol{z}_{t-1}}\frac{\partial \boldsymbol{z}_{t-1}}{\partial \boldsymbol{\epsilon }_t}\frac{\boldsymbol{\epsilon }_{t}^{c}\cdot\exp \left( \boldsymbol{Coe}_{t}^{\text{text}}+\boldsymbol{Coe}_{t}^{\text{layout}} \right)}{\left( \exp \left( \boldsymbol{Coe}_{t}^{\text{text}} \right) +\exp \left( \boldsymbol{Coe}_{t}^{\text{layout}} \right) \right) ^2}\right]\\=&\sum_b\left[\frac{\partial \mathcal{L}_b \left( \mathcal{A} _{t-1}^{\text{text}},\mathcal{A} _{t-1}^{\text{layout}} \right)}{\partial \mathcal{A} _{(j_b,t-1)}^{c}}\frac{\partial \mathcal{A} _{(j_b,t-1)}^{c}}{\partial \boldsymbol{z}_{t-1}}\left( \sqrt{1-\bar{\alpha}_{t-1}-\sigma ^2}-\!\frac{\sqrt{1-\bar{\alpha}_t}}{\sqrt{\alpha _t}} \right) \right.\\
&\left.\times\frac{\boldsymbol{\epsilon }_{t}^{c}\cdot\exp \left( \boldsymbol{Coe}_{t}^{\text{text}}+\boldsymbol{Coe}_{t}^{\text{layout}} \right)}{\left( \exp \left( \boldsymbol{Coe}_{t}^{\text{text}} \right) +\exp \left( \boldsymbol{Coe}_{t}^{\text{layout}} \right) \right) ^2}\right]\\
\end{aligned}
\end{equation}

% \\=&\sum_b{\frac{\left( \sum_i{\mathcal{M}_b}-1 \right) \mathcal{J}}{\left( \sum_i{\mathcal{A}_{\left( ij_b,t-1 \right)}^{\text{text}}} \right) ^2}}\frac{\partial \mathcal{A} _{t-1}^{\text{text}}}{\partial \boldsymbol{z}_{t-1}}\left( \sqrt{1-\bar{\alpha}_{t-1}-\sigma ^2}-\frac{\sqrt{1-\bar{\alpha}_t}}{\sqrt{\alpha _t}} \right) \frac{\boldsymbol{\epsilon }_{t}^{\text{text}}\cdot\exp \left( \boldsymbol{Coe}_{t}^{\text{text}}+\boldsymbol{Coe}_{t}^{\text{layout}} \right)}{\left( \exp \left( \boldsymbol{Coe}_{t}^{\text{text}} \right) +\exp \left( \boldsymbol{Coe}_{t}^{\text{layout}} \right) \right) ^2}

For any T2I and L2I models, we have the following:
\begin{equation}
\label{eq16}
\frac{\partial \mathcal{L}_b \left( \mathcal{A} _{t-1}^{\text{text}},\mathcal{A} _{t-1}^{\text{layout}} \right)}{\partial \mathcal{A} _{(j_b,t-1)}^{c}}=\frac{\mathcal{J} \sum_i{\left( \mathcal{A} _{\left( ij_b,t-1 \right)}^{c}\odot \mathcal{M}_b \right)}-\mathcal{M}_b\sum_i{\mathcal{A} _{\left( ij_b,t-1 \right)}^{c}}}{\left( \sum_i{\mathcal{A} _{\left( ij_b,t-1 \right)}^{c}} \right) ^2}
\end{equation}
where $\mathcal{J}$ is a matrix with all elements equal to $1$. All variables in Eq. \ref{eq14} are known, indicating the existence of the gradient in Eq. \ref{eq11}.

When using the loss function given by Eq. \ref{eq13} under any spatial-aware conditions, the gradient in Eq. \ref{eq11} can be derived as follows:
\begin{equation}
\begin{aligned}
\label{eq17}
&\frac{\partial \mathcal{L} \left( \mathcal{A} _{t-1}^{\text{text}},\mathcal{A} _{t-1}^{\text{spatial}} \right)}{\partial \boldsymbol{Coe}_{t}^{c}}\\=&\sum_b\left[\frac{\partial \mathcal{L}_b \left( \mathcal{A} _{t-1}^{\text{text}},\mathcal{A} _{t-1}^{\text{spatial}} \right)}{\partial \mathcal{A} _{(j_b,t-1)}^{c}}\frac{\partial \mathcal{A} _{(j_b,t-1)}^{c}}{\partial \boldsymbol{z}_{t-1}}\frac{\partial \boldsymbol{z}_{t-1}}{\partial \boldsymbol{\epsilon }_t}\frac{\partial \boldsymbol{\epsilon }_t}{\partial \boldsymbol{\xi }^c_t}\frac{\partial \boldsymbol{\xi }^c_t}{\partial \boldsymbol{Coe}_{t}^{c}}\right] \\=&\sum_b\left[\frac{\partial \mathcal{L}_b \left( \mathcal{A} _{t-1}^{\text{text}},\mathcal{A} _{t-1}^{\text{spatial}} \right)}{\partial \mathcal{A} _{(j_b,t-1)}^{c}}\frac{\partial \mathcal{A} _{(j_b,t-1)}^{c}}{\partial \boldsymbol{z}_{t-1}}\left( \sqrt{1-\bar{\alpha}_{t-1}-\sigma ^2}-\!\frac{\sqrt{1-\bar{\alpha}_t}}{\sqrt{\alpha _t}} \right) \right.\\
&\left.\times\frac{\boldsymbol{\epsilon }_{t}^{c}\cdot\exp \left( \boldsymbol{Coe}_{t}^{\text{text}}+\boldsymbol{Coe}_{t}^{\text{spatial}} \right)}{\left( \exp \left( \boldsymbol{Coe}_{t}^{\text{text}} \right) +\exp \left( \boldsymbol{Coe}_{t}^{\text{spatial}} \right) \right) ^2}\right]\\
\end{aligned}
\end{equation}
\begin{equation}
\label{eq18}
\frac{\partial \mathcal{L}_b \left( \mathcal{A} _{t-1}^{\text{text}},\mathcal{A} _{t-1}^{\text{spatial}} \right)}{\partial \mathcal{A} _{(j_b,t-1)}^{c}}=\frac{\mathcal{J} \sum_i{\left( \mathcal{A} _{\left( ij_b,t-1 \right)}^{c}\odot f_b(\mathcal{C}) \right)}-f_b(\mathcal{C})\sum_i{\mathcal{A} _{\left( ij_b,t-1 \right)}^{c}}}{\left( \sum_i{\mathcal{A} _{\left( ij_b,t-1 \right)}^{c}} \right) ^2}
\end{equation}
where $c\in\{\text{text}, \text{spatial}\}$.

Therefore, the gradient in Eq. \ref{eq11} exists for the selection of different loss functions.

%%%%%%%%%%%%%%%%%%%%%%%%%%%%%%%%%%%%%%%%%%%%%%%%%%%%%%%%%%%%%%%%%%%%%%%%%%%%%%%
%%%%%%%%%%%%%%%%%%%%%%%%%%%%%%%%%%%%%%%%%%%%%%%%%%%%%%%%%%%%%%%%%%%%%%%%%%%%%%%

\subsection{Inference details}\label{inference details}
We provide a detailed compositional denoising process for RealCompo, which achieves a complementary balance between the advantages of the T2I model and the spatial-aware diffusion model by combining their predicted noise during the denoising stage. We provide the pseudocode for the compositional denoising process of the layout-based RealCompo as followed, we have highlighted the innovations of our method in \textcolor{mycolor}{blue}.
\begin{algorithm}
   \caption{Compositional denoising procedure of layout-based RealCompo}
   \label{algorithm1}
    \begin{algorithmic}[1]
   \Statex {\bfseries Input:}  A text prompt $\mathcal{P}$, a set of layout $\mathcal{B}$, \textcolor{mycolor}{a pretrained T2I model and a pretrained L2I model} 
   \Statex {\bfseries Output:}  A clear latent $\boldsymbol z_0$ 
   \State $\boldsymbol z_T \sim \mathcal{N} (\mathbf{0},\mathbf{I}) $
   \State $\boldsymbol{Coe}^{\text{text}}_T=\boldsymbol{Coe}^{\text{layout}}_T\sim \mathcal{N}( \mathbf{0},\mathbf{I})$
		\For {$t=T,\ldots ,1$}
            \If{$t>t_0$}
            \State $\boldsymbol\epsilon_{t}, \_=\text{L2I}\left( \boldsymbol z_t, \mathcal{P}, \mathcal{B}, t \right) $
            \Else
				% \State $\boldsymbol\epsilon_1 \sim \mathcal{N} (\mathbf{0},\mathbf{I}) $ if $t>1$, else $\boldsymbol\epsilon_1 =\mathbf{0}$
                \State $\boldsymbol\epsilon_{t}^{\text{text}}, \_=\text{T2I}\left( \boldsymbol z_t, \mathcal{P}, t \right) $
				\State $\boldsymbol\epsilon_{t}^{\text{layout}}, \_=\text{L2I}\left( \boldsymbol z_t, \mathcal{P}, \mathcal{B}, t \right) $
                \State \textcolor{mycolor}{Get the balanced noise $\boldsymbol{\epsilon_t}$ from Eq. \ref{eq8} and Eq. \ref{eq9}}
                \State Get the denoised latent $\boldsymbol{z}_{t-1}$ from Eq. \ref{eq3}
                \State $\boldsymbol\epsilon_{t-1}^{\text{text}}, \mathcal{A}_{t-1}^{\text{text}}=\text{T2I}\left( \boldsymbol z_{t-1}, \mathcal{P}, t \right) $
                \State $\boldsymbol\epsilon_{t-1}^{\text{layout}}, \mathcal{A}_{t-1}^{\text{layout}}=\text{L2I}\left( \boldsymbol z_{t-1}, \mathcal{P}, \mathcal{B}, t \right) $
                \State \textcolor{mycolor}{Compute $\mathcal{L}(\mathcal{A}_{t-1}^{\text{text}}, \mathcal{A}_{t-1}^{\text{layout}})$ from Eq. \ref{eq10} }
			  \State \textcolor{mycolor}{Update $\boldsymbol{Coe}_t^c$ according to Eq. \ref{eq11}}
                \State \textcolor{mycolor}{Get the balanced noise $\boldsymbol{\epsilon_t}$ from Eq. \ref{eq8} and Eq. \ref{eq9}}
            \EndIf
            \State Get the denoised latent $\boldsymbol{z}_{t-1}$ from Eq. \ref{eq3}
		\EndFor
            \State \textbf{return} $\boldsymbol z_0$
        \end{algorithmic}
\end{algorithm}

\newpage
\subsection{Gradient Analysis}\label{gradient analysis}

\begin{figure}[ht!]
\vskip -0.1in
\begin{center}
\centerline{\includegraphics[width=.65\textwidth]{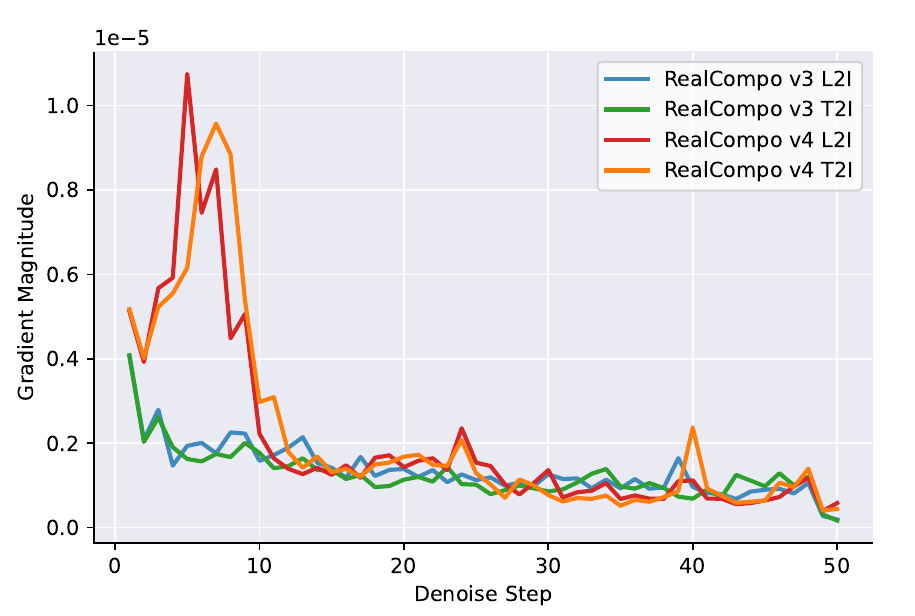}}
\vskip -0.1in
\caption{Changes of gradient magnitude in Eq. \ref{eq11} across all denoising process for the T2I and L2I models of RealCompo v3 and v4.}
\label{gradient}
\end{center}
\vskip -0.3in
\end{figure}

\paragraph{Gradient Analysis}
We selected RealCompo v3 and v4 to analyze the gradient changes in Eq. \ref{eq11} across all denoising stages. As shown in Fig. \ref{gradient}, we use the same prompt and random seed to visualize the gradient magnitude changes corresponding to T2I and L2I for each model version. We observe that the gradient magnitude change of RealCompo v4 fluctuated more in the early denoising stages. We argue that TokenCompose, which enhances the composition capability of multiple-object generation by fine-tuning the model using segmentation masks, may overlap in functionality with the layout-based multiple-object generation, and TokenCompose's positioning of objects may not consistently align with the bounding box. Therefore, RealCompo must focus on balancing the positioning of TokenCompose and layout in the early denoising stages, leading to less stable gradients compared to RealCompo v3. Additionally, due to LayGuide's weaker positioning ability compared to GLIGEN, RealCompo v4 may occasionally generate objects with less coverage of the bounding box, as mentioned in the ablation experiment in \cref{ablation exp}.

\subsection{Limitations and Future Work}
\label{Limitations}
\paragraph{Limitations}While our RealCompo enhances both realism and compositionality in a training-free manner, it should be noted that the computational cost of our method is slightly higher compared to that of a single T2I model or a single spatial-aware image diffusion model, due to the need to combine two models and compute loss and gradients. However, by adjusting the combination stage of RealCompo, we can keep the computational cost within an acceptable range.

\paragraph{Future Work}In future work, we aim to explore more efficient computational methods to improve the calculation efficiency of RealCompo while maintaining high-quality results. Additionally, we plan to extend its application to more challenging tasks such as text-to-video and text-to-3D generation.

\subsection{Broader Impact}\label{broader impact}Recent significant advancements in text-to-image diffusion models have opened up new possibilities for creative design, autonomous media, and various other sectors. However, the dual-use nature of this technology raises concerns about its social impact. Image diffusion models carry the risk of misuse, particularly in the realm of impersonating humans. For example, in today's society, malicious applications such as "deepfakes" have been employed in inappropriate contexts to fabricate attacks on specific public figures. It is crucial to clarify that our algorithm is designed to enhance the quality of image generation, and we do not endorse or facilitate such malicious applications.

% \paragraph{Broader Impact}
% This paper presents work whose goal is to advance the field of Machine Learning. There are many potential societal consequences of our work, none of which we feel must be specifically highlighted here.

\newpage
\section{More Generation Results}\label{more results}

\begin{figure*}[ht!]
\vskip 0.2in
\begin{center}
\centerline{\includegraphics[width=.99\textwidth]{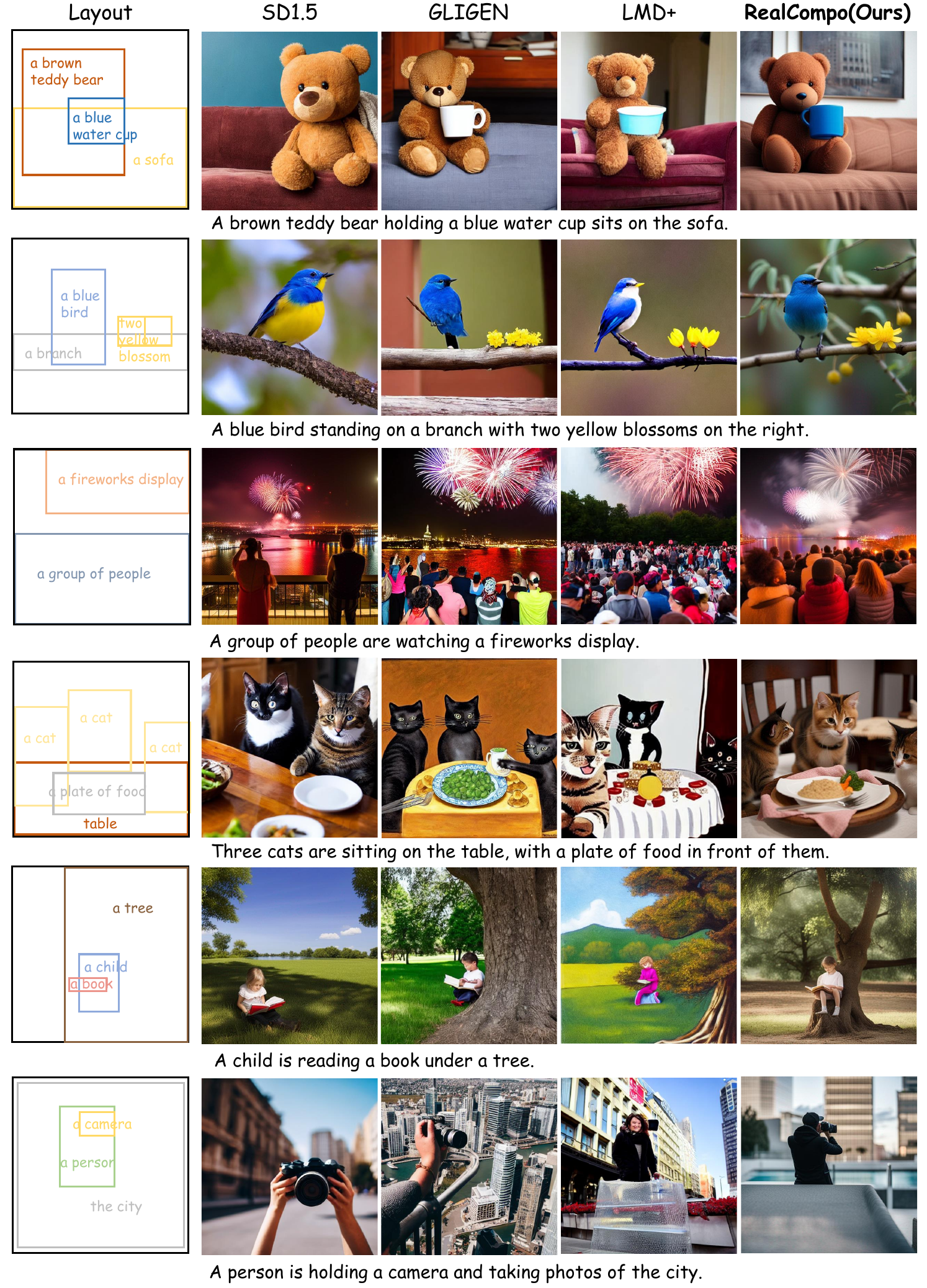}}
\caption{More generation results about layout-based RealCompo.}
\label{appendix}
\end{center}
\vskip -0.2in
\end{figure*}

\begin{figure*}[ht!]
\vskip -0.05in
\begin{center}
\centerline{\includegraphics[width=.86\textwidth]{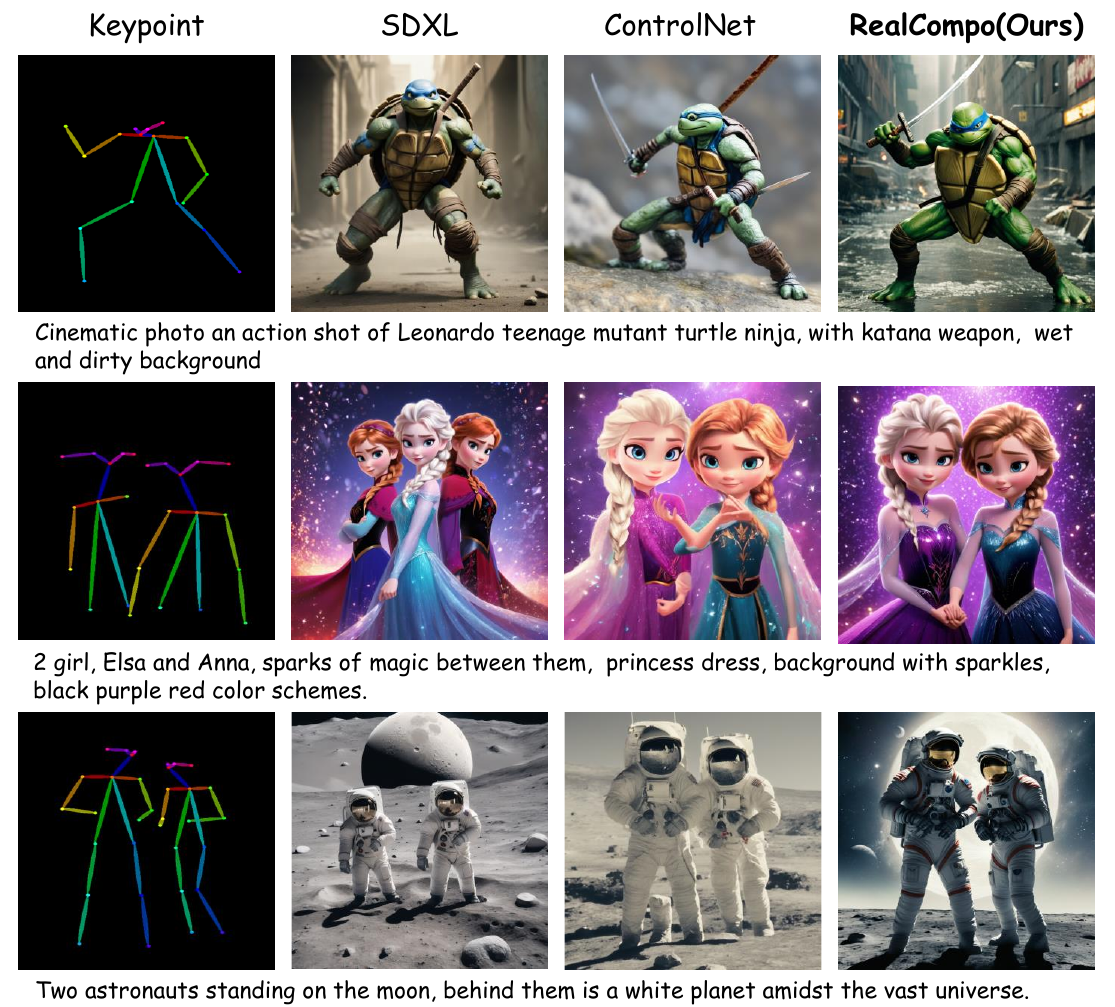}}
\vskip -0.1in
\caption{More generation results about keypoint-based RealCompo.}
\label{appendix_key}
\end{center}
\vskip -0.2in
\end{figure*}

\begin{figure*}[ht!]
\vskip -0.05in
\begin{center}
\centerline{\includegraphics[width=.86\textwidth]{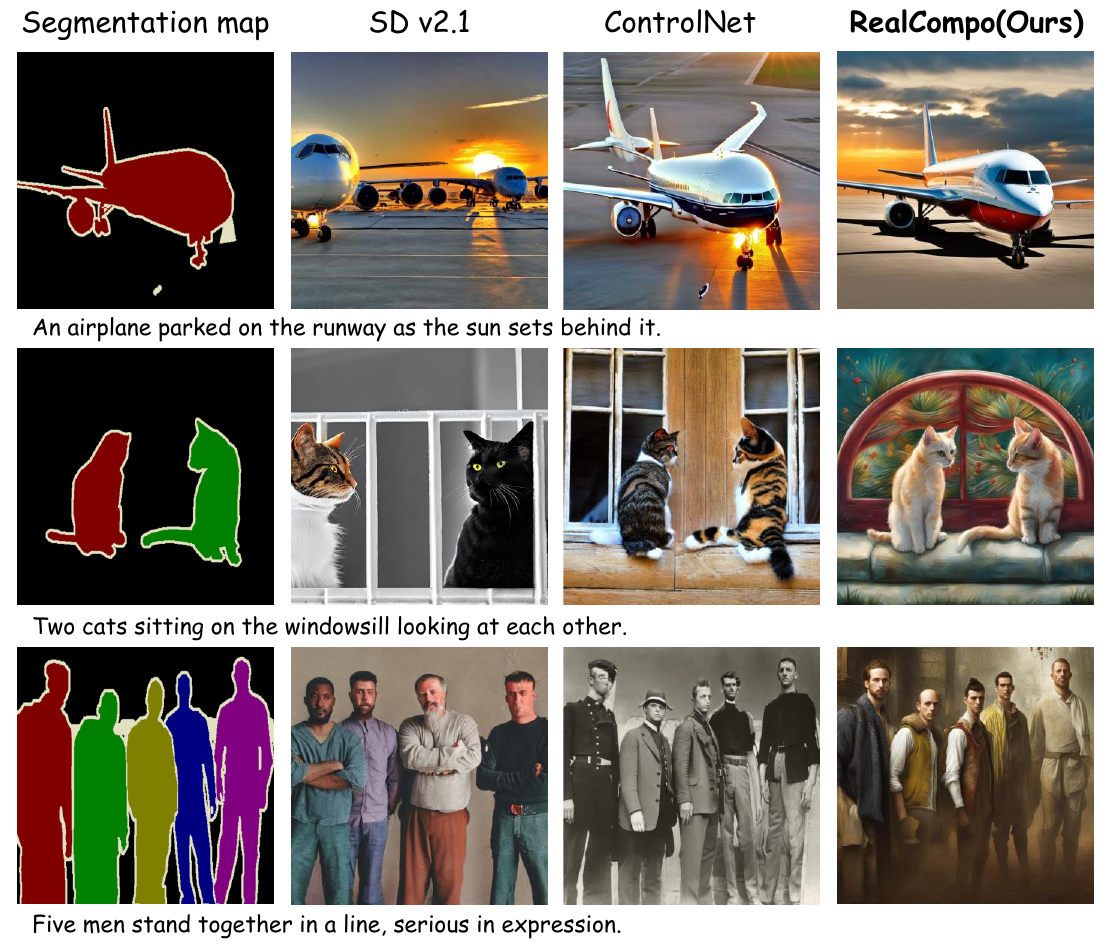}}
\vskip -0.1in
\caption{More generation results about segmentation-based RealCompo.}
\label{appendix_seg}
\end{center}
\vskip -0.2in
\end{figure*}

\end{document}